\documentclass[conference]{IEEEtran}
\IEEEoverridecommandlockouts

\newif\ifarxiv
\arxivtrue

\newif\iffinal
\finaltrue

\newif\ifieee
\ieeefalse

\usepackage{cite}
\usepackage{amsmath,amssymb,amsfonts}
\usepackage{algorithmic}
\usepackage{graphicx}
\usepackage{textcomp}

\usepackage[caption=false,farskip=0pt]{subfig}
\usepackage{booktabs}
\usepackage{multirow}
\usepackage[hyphens]{url}
\usepackage{balance}

\ifieee
\else
\usepackage[colorlinks=true,allcolors=black]{hyperref} %
\fi

\usepackage[acronym]{glossaries}
\usepackage{xspace}
\usepackage[utf8]{inputenc}
\usepackage[T1]{fontenc}
\usepackage[dvipsnames,table]{xcolor}
\usepackage{diagbox}
\usepackage[capitalise]{cleveref} %
\usepackage{comment}

\usepackage{soul}

\newcounter{fncounter}
\newcommand\customfootnote[1]{\stepcounter{fncounter}\footnote{\hspace{0.2mm}#1}}

\newcommand{\customIndent}{\hspace{2mm}}

\newcommand{\figTeaser}{Fig.~1\xspace}

\newcommand*{\RL}[2][]{\textcolor{Rhodamine}{[\textbf{\ifthenelse{\equal{#1}{}}{RL}{RL(#1)}}: #2]}}
\newcommand\RLI[1]{} %
\newcommand*{\DM}[2][]{\textcolor{blue}{[\textbf{\ifthenelse{\equal{#1}{}}{DM}{DM(#1)}}: #2]}}
\newcommand*{\VE}[2][]{\textcolor{ForestGreen}{[\textbf{\ifthenelse{\equal{#1}{}}{VE}{VE(#1)}}: #2]}}

\newacronymstyle{long-short-br}
{%
  \GlsUseAcrEntryDispStyle{long-short}%
}%
{%
  \GlsUseAcrStyleDefs{long-short}%
}
\setacronymstyle{long-short-br}

\newcommand{\teaserfigure}{
    \begin{center}
    
    \resizebox{0.985\linewidth}{!}{
    \begin{tabular}{@{}
                m{0.025\linewidth} @{}
                | @{\hspace{0.0075\linewidth}}
                >{\centering}m{0.1\linewidth} @{\hspace{0.0075\linewidth}}
                >{\centering}m{0.1\linewidth} @{\hspace{0.0075\linewidth}}
                >{\centering}m{0.1\linewidth} @{\hspace{0.0115\linewidth}}
                 @{\hspace{0.0075\linewidth}}
                >{\centering}m{0.1\linewidth} @{\hspace{0.0075\linewidth}}
                >{\centering}m{0.1\linewidth} @{\hspace{0.0075\linewidth}}
                >{\centering}m{0.1\linewidth} @{\hspace{0.0115\linewidth}}
                 @{\hspace{0.0075\linewidth}}
                >{\centering}m{0.1\linewidth} @{\hspace{0.0075\linewidth}}
                >{\centering}m{0.1\linewidth} @{\hspace{0.0075\linewidth}} >{\centering\arraybackslash}m{0.1\linewidth} @{} }
            & \multicolumn{3}{c@{\hspace{0.0075\linewidth}}@{\hspace{0.0075\linewidth}}}{AOLP~\cite{hsu2013application} (Protocol A)}
            & \multicolumn{3}{c@{\hspace{0.0075\linewidth}}@{\hspace{0.0075\linewidth}}}{AOLP~\cite{hsu2013application} (Protocol B)}
            & \multicolumn{3}{c}{CCPD~\cite{xu2018towards} (latest version)} \\[-0.85ex]
            \rotatebox{90}{\small \centering Training} &
            \includegraphics[width=\linewidth]{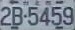} &
            \includegraphics[width=\linewidth]{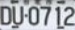} &
            \includegraphics[width=\linewidth]{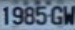} &
            \includegraphics[width=\linewidth]{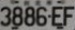} &
            \includegraphics[width=\linewidth]{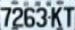} &
            \includegraphics[width=\linewidth]{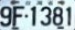} &
            \includegraphics[width=\linewidth]{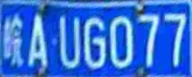} &
            \includegraphics[width=\linewidth]{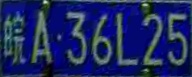} &
            \includegraphics[width=\linewidth]{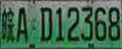} \\[1ex]
            \rotatebox{90}{\small \centering Test} &
            \includegraphics[width=\linewidth]{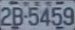} &
            \includegraphics[width=\linewidth]{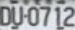} &
            \includegraphics[width=\linewidth]{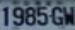} &
            \includegraphics[width=\linewidth]{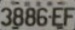} &
            \includegraphics[width=\linewidth]{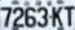} &
            \includegraphics[width=\linewidth]{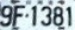} &
            \includegraphics[width=\linewidth]{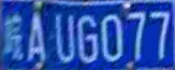} &
            \includegraphics[width=\linewidth]{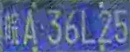} &
            \includegraphics[width=\linewidth]{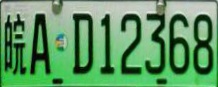} \\           
        \end{tabular}%
    } \quad
    
    \vspace{-0.75mm}
    \end{center}
    
    {\footnotesize Fig. 1.~Examples of near-duplicates in the training and test sets of the \aolp~\cite{hsu2013application} and \ccpd~\cite{xu2018towards} datasets, which are by far the two most popular datasets in the \acrfull*{lpr} literature.
    The top row shows license plates cropped and rectified from images in the training sets, while the bottom row shows license plates cropped and rectified from their nearest neighbors in the respective test set. We show three image pairs for each dataset representing the $10$th, $50$th and $90$th percentiles based on their Euclidean distance in pixel space. Protocols A and B in the AOLP dataset are described in the main text.}
    \label{fig:teaser}

}

\makeatletter
\apptocmd\@maketitle{{\vspace{1\baselineskip}\teaserfigure{}\par}}{}{}
\makeatother

\let\OLDthebibliography\thebibliography
\renewcommand\thebibliography[1]{
  \OLDthebibliography{#1}
  \setlength{\itemsep}{2pt}
}

\usepackage{transparent}
\usepackage{tikz}
\ifarxiv
\newcommand\copyrighttext{%
  \scriptsize Accepted at IJCNN 2023. The final published version is available on IEEE Xplore (DOI: \href{https://doi.org/10.1109/IJCNN54540.2023.10191584}{\textcolor{blue}{10.1109/IJCNN54540.2023.10191584}}).}
\newcommand\copyrightnotice{%
\begin{tikzpicture}[remember picture,overlay]
\node[anchor=south,yshift=30pt,xshift=0pt] at (current page.south) {\fbox{\transparent{0.85}\parbox{\dimexpr0.72\textwidth-\fboxsep-\fboxrule\relax}{\copyrighttext}}};
\end{tikzpicture}%
}
\else
\fi

\begin{document}

\newacronym{ac}{AC}{access control}
\newacronym{alpr}{ALPR}{Automatic License Plate Recognition}
\newacronym{ctc}{CTC}{Connectionist Temporal Classification}
\newacronym{le}{LE}{traffic law enforcement}
\newacronym{lp}{LP}{license plate}
\newacronym{lpd}{LPD}{License Plate Detection}
\newacronym{lpr}{LPR}{License Plate Recognition}
\newacronym{ocr}{OCR}{Optical Character Recognition}
\newacronym{rp}{RP}{road patrol}
\newacronym{starnet}{STAR-Net}{SpaTial Attention Residue Network}
\newacronym{trba}{TRBA}{TPS-ResNet-BiLSTM-Attention}

\newcommand{\aolp}{AOLP\xspace}
\newcommand{\ccpd}{CCPD\xspace}
\newcommand{\ccpdfair}{CCPD-Fair\xspace}
\newcommand{\chineselp}{ChineseLP\xspace}
\newcommand{\clpd}{CLPD\xspace}
\newcommand{\englishlp}{EnglishLP\xspace}
\newcommand{\medialab}{Medialab LPR\xspace}
\newcommand{\pku}{PKU\xspace}
\newcommand{\reid}{ReId\xspace}

\newcommand{\crnn}{CRNN\xspace}
\newcommand{\starnet}{\acrshort*{starnet}\xspace}
\newcommand{\trba}{\acrshort*{trba}\xspace}
\newcommand{\vitstrbase}{ViTSTR-Base\xspace}
\newcommand{\vitstrsmall}{ViTSTR-Small\xspace}
\newcommand{\vitstrtiny}{ViTSTR-Tiny\xspace}
\newcommand{\vitstr}{ViTSTR\xspace}

\newcommand{\holistic}{Holistic-CNN\xspace}
\newcommand{\multitask}{Multi-Task\xspace}
\newcommand{\multitaskLR}{Multi-Task-LR\xspace}
\newcommand{\cnng}{CNNG\xspace}
\iffinal
\newcommand{\supplementary}{\url{https://raysonlaroca.github.io/supp/lpr-train-on-test/}}
\newcommand{\supplementaryEndParagraph}{\supplementary}
\else
\newcommand{\supplementary}{[\textit{hidden for review}]}
\newcommand{\supplementaryEndParagraph}{[\textit{hidden for~review}]}
\fi

\iffinal
\title{Do We Train on Test Data? The Impact of Near-Duplicates on License Plate~Recognition
}
\else
\title{Do We Train on Test Data? The Impact of Near-Duplicates on License Plate~Recognition
\thanks{The funding agencies are hidden for review.}
}
\fi

\iffinal
\author{
\begin{tabular}{cc}
\multicolumn{2}{c}{
Rayson Laroca\IEEEauthorrefmark{1}, Valter Estevam\IEEEauthorrefmark{1}$^,$\IEEEauthorrefmark{2}, Alceu S. Britto Jr.\IEEEauthorrefmark{3}, Rodrigo Minetto\IEEEauthorrefmark{4}, and David Menotti\IEEEauthorrefmark{1}} \\[1ex] \small
\quad\IEEEauthorrefmark{1}\hspace{0.15mm}Federal University of Paran\'a, Curitiba, Brazil\quad & \small \quad\IEEEauthorrefmark{2}\hspace{0.15mm}Federal Institute of Paran\'a, Irati, Brazil\quad \\[-2.5pt]
\small \quad\IEEEauthorrefmark{3}\hspace{0.15mm}Pontifical Catholic University of Paran\'a, Curitiba, Brazil\quad & \small \quad\IEEEauthorrefmark{4}\hspace{0.15mm}Federal University of Technology-Paran\'a, Curitiba, Brazil\quad \\
\multicolumn{2}{c}{\resizebox{0.875\linewidth}{!}{
\hspace{3mm}\IEEEauthorrefmark{1}{\hspace{-0.45mm}\tt\small \{rblsantos,vlejunior,menotti\}@inf.ufpr.br} \quad \IEEEauthorrefmark{3}{\hspace{0.15mm}\tt\small alceu@ppgia.pucpr.br} \quad \IEEEauthorrefmark{4}{\hspace{0.15mm}\tt\small rminetto@utfpr.edu.br}
}}
\end{tabular}
}
\else
\author{\IEEEauthorblockN{Anonymous Authors}}
\fi

\maketitle

\copyrightnotice

\glsresetall
\begin{abstract}
This work draws attention to the large fraction of near-duplicates in the training and test sets of datasets widely adopted in \gls*{lpr} research.
These duplicates refer to images that, although different, show the same license plate.
Our experiments, conducted on the two most popular datasets in the field, show a substantial decrease in recognition rate when six well-known models are trained and tested under fair splits, that is, in the absence of duplicates in the training and test sets.
Moreover, in one of the datasets, the ranking of models changed considerably when they were trained and tested under duplicate-free splits.
These findings suggest that such duplicates have significantly biased the evaluation and development of deep learning-based models for \gls*{lpr}.
The list of near-duplicates we have found and proposals for fair splits are publicly available for further research at {\supplementaryEndParagraph}.
\end{abstract}

\section{Introduction}
\label{sec:introduction}

\glsresetall

Research into \gls*{alpr} has gained significant attention in recent years due to its practical applications, including toll collection, vehicle access control in restricted areas, and traffic law enforcement~\cite{zhuang2018towards,weihong2020research,wang2022rethinking}.

\gls*{alpr} is commonly divided into two tasks: \gls*{lpd} and \gls*{lpr}.
The first task refers to locating the \glspl*{lp} in the input image, while the second refers to recognizing the characters on those \glspl*{lp}.
Recent developments in deep neural networks have led to advancements in both tasks, but current research has mostly focused on \gls*{lpr}~\cite{laroca2022cross,zhang2021robust_attentional,liu2021fast,zhang2021efficient,nascimento2022combining} since general-purpose object detectors (e.g., Faster-RCNN and YOLO) have already achieved notable success in \gls*{lpd} for some time now~\cite{hsu2017robust,laroca2018robust,zhang2018joint}.

\gls*{lpr} methods are typically evaluated using images from public datasets, which are divided into disjoint training and test sets using standard splits, defined by the datasets' authors, or following previous works (when there is no standard split).
In most cases, such an assessment is carried out independently for each dataset~\cite{laroca2018robust,zhuang2018towards,weihong2020research,zhang2021efficient,pham2022effective}.

Although the images for training and testing belong to disjoint sets, the splits traditionally adopted in the literature were defined without considering that the same \gls*{lp} may appear in multiple images (see \cref{sec:datasets-duplicates}).
As a result, we found that there are many \emph{near-duplicates} (i.e., different images of the same \gls*{lp}) in the training and test sets of datasets widely explored in \gls*{alpr} research.
In this study, to evaluate the impact of such duplicates on \gls*{lpr}, we focus our analysis on the \aolp~\cite{hsu2013application} and \ccpd~\cite{xu2018towards} datasets, as they are the most popular datasets in the field.
Nevertheless, \cref{sec:other-datasets} clarifies the existence of near-duplicates in several other datasets and gives examples of how it has been overlooked in the~literature.

Considering that recent \gls*{alpr} approaches rectify (unwarp) the detected \glspl*{lp} before feeding them to the recognition model~\cite{fan2022improving,qin2022towards,silva2022flexible,wang2022rethinking,xu2022eilpr}, the presence of duplicates in the training and test sets means that \gls*{lpr} models are, in many cases, being trained and tested on essentially the same images (see \figTeaser).
This is a critical issue for accurate scientific evaluation~\cite{barz2020do,emami2020analysis}.
Researchers aim to compare models in terms of their ability to generalize to unseen data~\cite{feldman2020what,liao2021are}.
With a considerable number of duplicates, however, there is a risk of comparing the models in terms of their ability to memorize training data, which increases with the model's capacity~\cite{barz2020do,hooker2020characterising}.

In light of this, we create \emph{fair splits} for the \aolp and \ccpd datasets (see \cref{sec:experiments-fair-datasets}) and compare the performance of six well-known \gls*{ocr} models applied to \gls*{lpr} under the original (adopted in previous works) and fair protocols.
Our results indicate that the presence of duplicates greatly affects the performance evaluation of these models.
Considering the experiments under the AOLP-B protocol as an example, the model that reached the best results under the traditional split ranked third under the fair one.
Such results imply that the duplicates have biased the evaluation and development of deep learning-based models for~\gls*{lpr}.

This work is inspired by~\cite{barz2020do}, where duplicates in the CIFAR-10 and CIFAR-100 datasets were identified, and motivated by recent studies that demonstrated the existence of bias in the \gls*{alpr} context.
An example worth mentioning is~\cite{laroca2022cross}, where the authors observed significant drops in \gls*{lpr} performance  when training and testing state-of-the-art models in a leave-one-dataset-out experimental setup.

\RLI{differente de outras áreas (como á area geral de reconhecimento de objetos), aqui é bastante fácil identificar imagens duplicadas (usando a placa como referencia). Entretanto, pesquisadores e criados de datasets não tem dado atenção para este problema.}

In summary, this paper has two main contributions:
\begin{itemize}
    \item We unveil the presence of near-duplicates in the training and test sets of datasets widely adopted in the \gls*{alpr} literature.
    Our analysis, using the \aolp and \ccpd datasets, shows the impact of such duplicates on the evaluation of six well-known \gls*{ocr} models applied to~\gls*{lpr}.
    \begin{itemize}
        \item Our results on the AOLP dataset indicate that the high fraction of near-duplicates in the splits traditionally employed in the literature may have hindered the development and acceptance of more efficient \gls*{lpr} models that have strong generalization abilities but do not memorize duplicates as well as other~models;
        \item Our experiments on the \ccpd dataset give a clearer picture of the true capabilities of \gls*{lpr} models compared to prior evaluations using the standard split, in which the test set has duplicates in the training set. Results revealed a decrease in the average recognition rate from $80.3$\% to $77.6$\% when the experiments were conducted under a fair split without~duplicates.
        
    \end{itemize}
    \item We create and release \emph{fair splits} for these datasets where there are no duplicates in the training and test sets, and the key characteristics of the original partitions are preserved as much as possible (see details on \cref{sec:experiments-fair-datasets}).
\end{itemize}

This paper is structured as follows.
We describe the \aolp and \ccpd datasets in \cref{sec:datasets}, detailing the protocols often adopted for each and how many near-duplicates they have.
\cref{sec:experiments} details the experiments performed.
The presence of duplicates in other popular datasets is discussed in \cref{sec:other-datasets}.
Finally, conclusions are provided in \cref{sec:conclusions}.
\section{The \aolp and \ccpd datasets}
\label{sec:datasets}

The two most popular datasets for \gls*{alpr} (in terms of the number of works that explored them) are \aolp~\cite{hsu2013application} and \ccpd~\cite{xu2018towards}.
While most authors explored at least one of these two datasets in their experiments (e.g.,~\cite{li2019toward,laroca2021efficient,silva2022flexible,wang2022rethinking}), there are many works in which the experiments were performed exclusively on them (e.g.,~\cite{xie2018new,zhang2020robust_license,liang2022egsanet,pham2022effective}).

\aolp was created to verify that \gls*{alpr} is better handled in an application-oriented way.
It is categorized into three subsets: \gls*{ac}, \gls*{le}, and \gls*{rp}.
The subsets have $681$, $757$ and $611$ images, respectively.
All images were acquired in the Taiwan~region.

As the \aolp dataset does not have a standard split, it has been divided in various ways in the literature.
For instance, some authors (e.g.,~\cite{xie2018new,laroca2021efficient,liang2022egsanet}) randomly divided its images into training and test sets with a $2$:$1$ ratio (we refer to this protocol as~\emph{AOLP-A}), whereas other authors (e.g.,~\cite{li2019toward,zhang2021efficient,wang2022rethinking}) used images from different subsets for training and testing, for example, the authors of \cite{fan2022improving,nguyen2022efficient,qin2022towards} used images from the \gls*{ac} and \gls*{le} subsets to train the proposed models and tested them on the \gls*{rp} subset (we refer to this protocol as~\emph{AOLP-B}).
Zhuang et al.~\cite{zhuang2018towards} evaluated their method under both the AOLP-A and AOLP-B protocols.
As commonly done in previous works, we consider that 20\% of the training images are allocated for validation in both protocols.

In 2018, Xu et al.~\cite{xu2018towards} claimed that the \gls*{alpr} datasets available at the time (including \aolp) either lacked quantity (i.e., they had less than $10$K images) or diversity (i.e., they were collected by static cameras or in overly controlled settings).
Thus, to assist in better benchmarking \gls*{alpr} approaches, they presented the \ccpd~dataset.

\ccpd comprises images taken with handheld cameras by workers of a roadside parking management company on the streets of a capital city in mainland China.
\newcounter{ccpdGit}
\setcounter{ccpdGit}{\thefncounter}
The dataset was updated/expanded twice after being introduced in 2018\customfootnote{CCPD's latest version: \url{https://github.com/detectRecog/CCPD/}}.
It originally consisted of $250$K images, divided into subsets (e.g., Blur, Challenge, Rotate, Weather, among others) according to their characteristics~\cite{xu2018towards}.
Then, in 2019, the authors released a new version --~much more challenging than the previous one~-- containing over $300$K images, refined annotations\customfootnote{While the annotations were refined in the first update to \ccpd (in 2019), there are still significant inaccuracies in the coordinates of the \gls*{lp}~corners.}, and a standard split protocol.
In summary, in this protocol, the $200$K images in the ``Base'' subset are split into training and validation sets (50\%/50\%), while all images from the other subsets are employed for testing.
Finally, in 2020, the authors included a new subset (Green) with $11{,}776$ images of electric vehicles, which have green \glspl*{lp} with eight characters (all the other subsets have images of vehicles with blue  \glspl*{lp} containing seven characters).
These updates to the \ccpd dataset are precisely why some works claim that it has $250$K images~\cite{zhang2020robust_license,fan2022improving,liang2022egsanet}, others claim that it has $280$-$290$K~\cite{zhang2021efficient,zhang2021robust_attentional,wang2022rethinking}, while the current version has $366{,}789$~images\footnotemark[\theccpdGit].

\subsection{Duplicates}
\label{sec:datasets-duplicates}

\begin{figure}[!htb]
    \centering
    \captionsetup[subfigure]{skip=1.5pt} 

    \resizebox{0.99\linewidth}{!}{
    \subfloat[][Subset AC]{
        \includegraphics[width=0.32\linewidth]{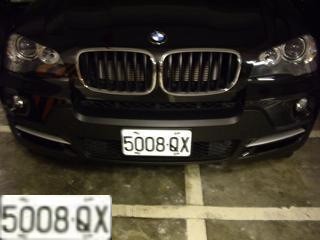}
    }
    \subfloat[][Subset LE]{
        \includegraphics[width=0.32\linewidth]{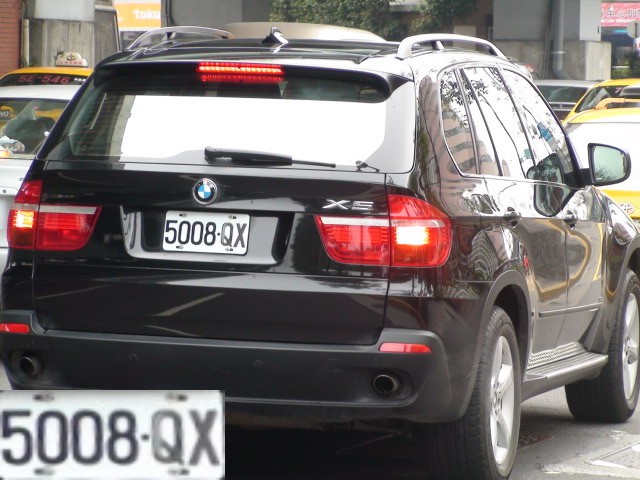}
    }
    \subfloat[][Subset RP]{
        \includegraphics[width=0.32\linewidth]{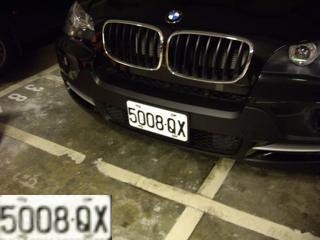}
    } \,
    }

    \vspace{2mm}
    
    \resizebox{0.99\linewidth}{!}{
    \subfloat[][Subset AC\label{fig:same-vehicle-examples-aolp-d}]{
        \includegraphics[width=0.32\linewidth]{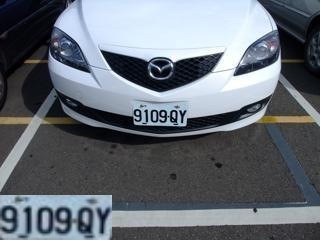}
    }
    \subfloat[\label{fig:same-vehicle-examples-aolp-e}][Subset AC]{
        \includegraphics[width=0.32\linewidth]{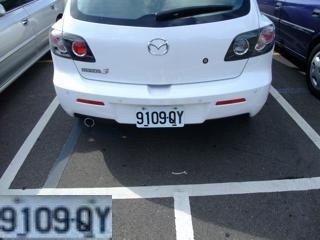}
    }
    \subfloat[][Subset RP]{
        \includegraphics[width=0.32\linewidth]{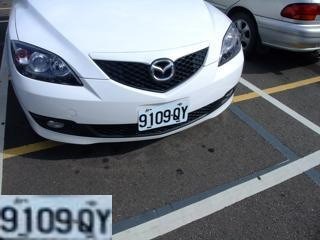}
    } \,
    }

    \vspace{-0.3mm}
    
    \caption{
    Examples of images from different subsets in the \aolp dataset~\cite{hsu2013application} that show the same vehicle/\gls*{lp}.
    In the split protocols often adopted in the literature (AOLP-A and AOLP-B), some of these images are in the training set and others are in the test set.
    We show a zoomed-in version of the rectified \gls*{lp} in the lower left region of each image for better~viewing.
    }
    \label{fig:same-vehicle-examples-aolp}
\end{figure}

\begin{figure}[!htb]
    \vspace{1mm}
    
    \centering
    \captionsetup[subfigure]{skip=-0.5pt,labelformat=empty,font=small}

    \resizebox{0.89\linewidth}{!}{
    \subfloat[][(a) Training set]{
    \captionsetup[subfigure]{skip=1pt,labelformat=empty,font=footnotesize}
    
    \subfloat[][Subset Base]{
    \includegraphics[width=0.24\linewidth]{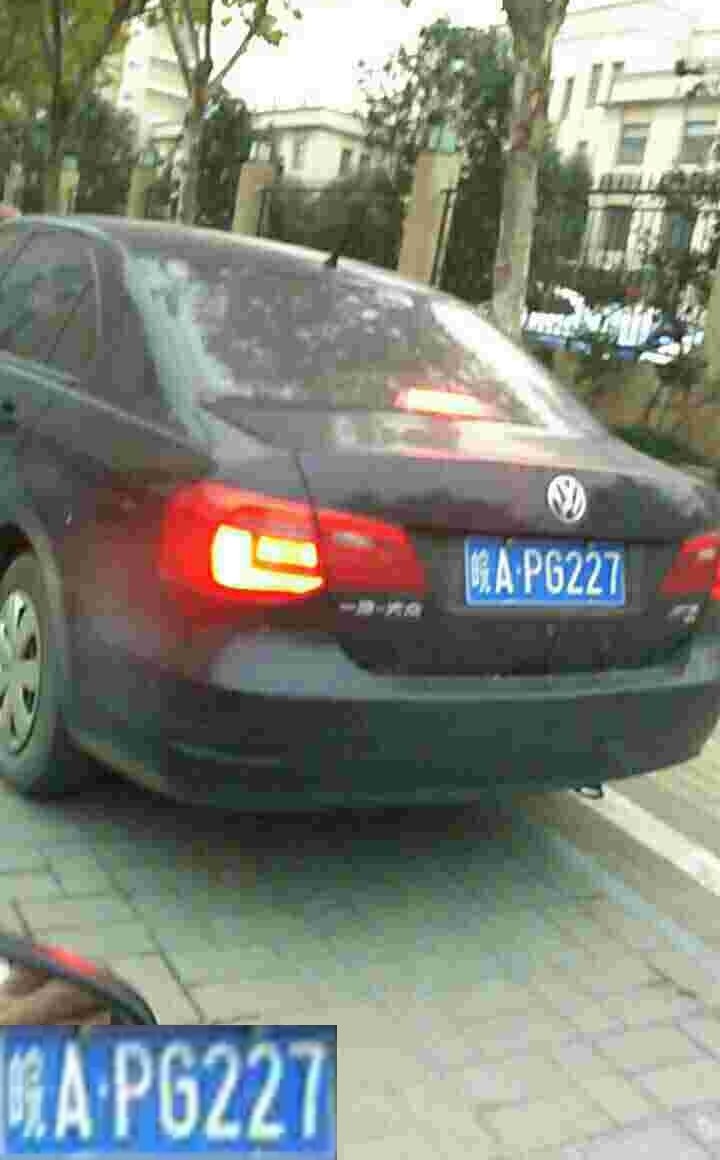}
    }
    
    \subfloat[][Subset Base]{
    \includegraphics[width=0.24\linewidth]{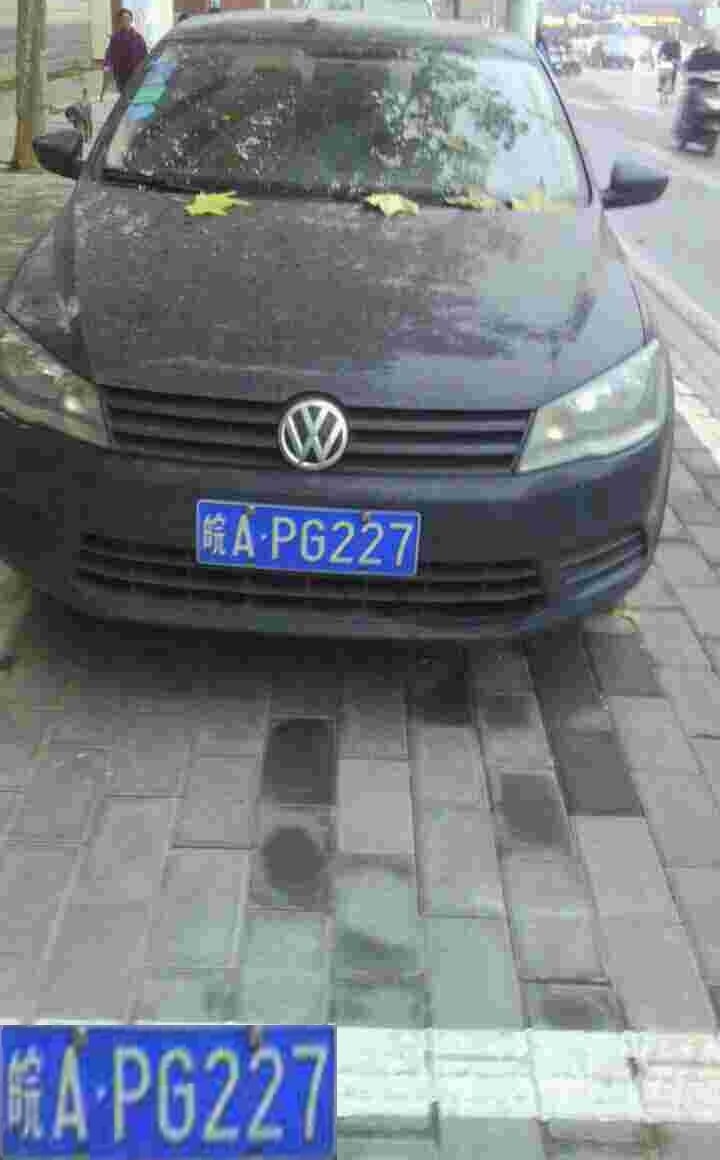}
    }
    
    \subfloat[][Subset Base]{
    \includegraphics[width=0.24\linewidth]{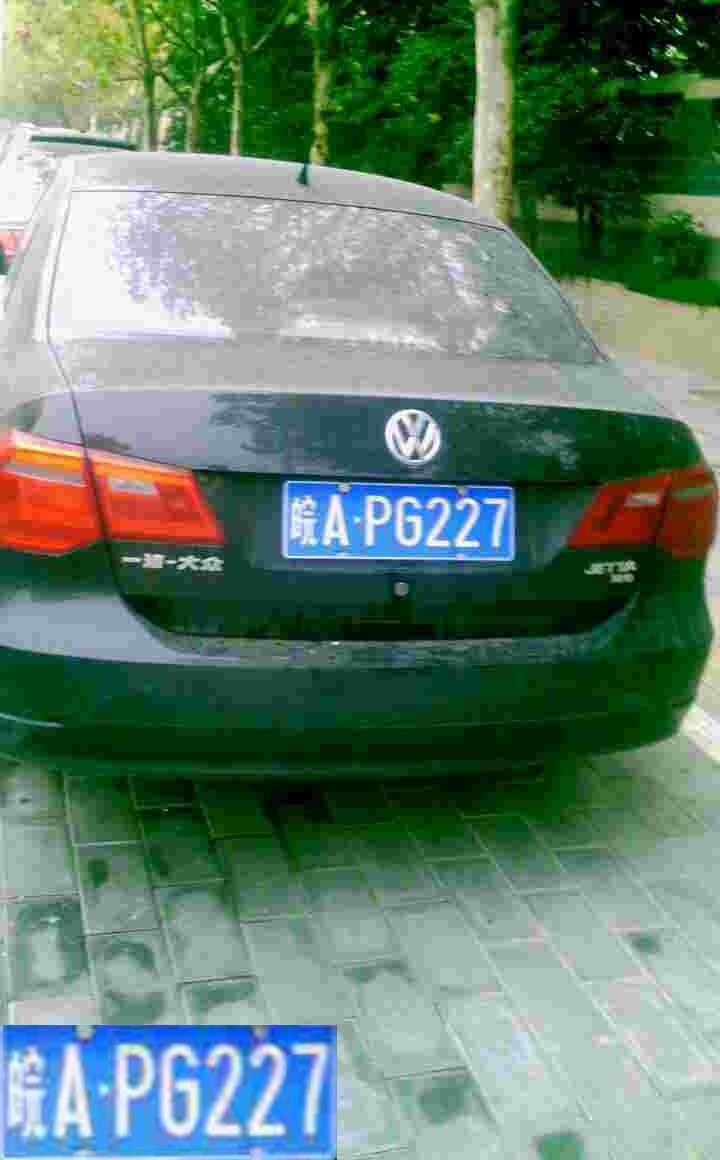}
    }
    
    \subfloat[][Subset Base]{
    \includegraphics[width=0.24\linewidth]{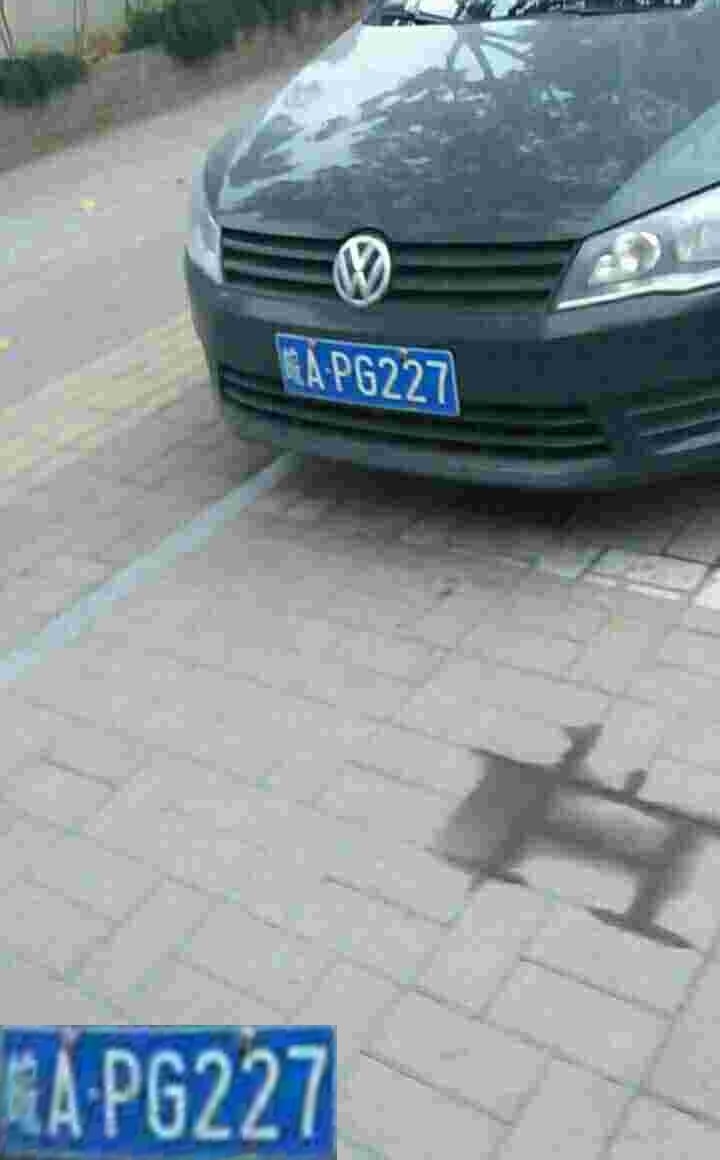}
    }
    }
    }

    \vspace{2.5mm}

    \setcounter{subfigure}{4}
    \resizebox{0.89\linewidth}{!}{
    \subfloat[][(b) Test set]{
    \captionsetup[subfigure]{skip=1pt,labelformat=empty,font=footnotesize}
    
    \subfloat[][Subset Challenge]{
    \includegraphics[width=0.24\linewidth]{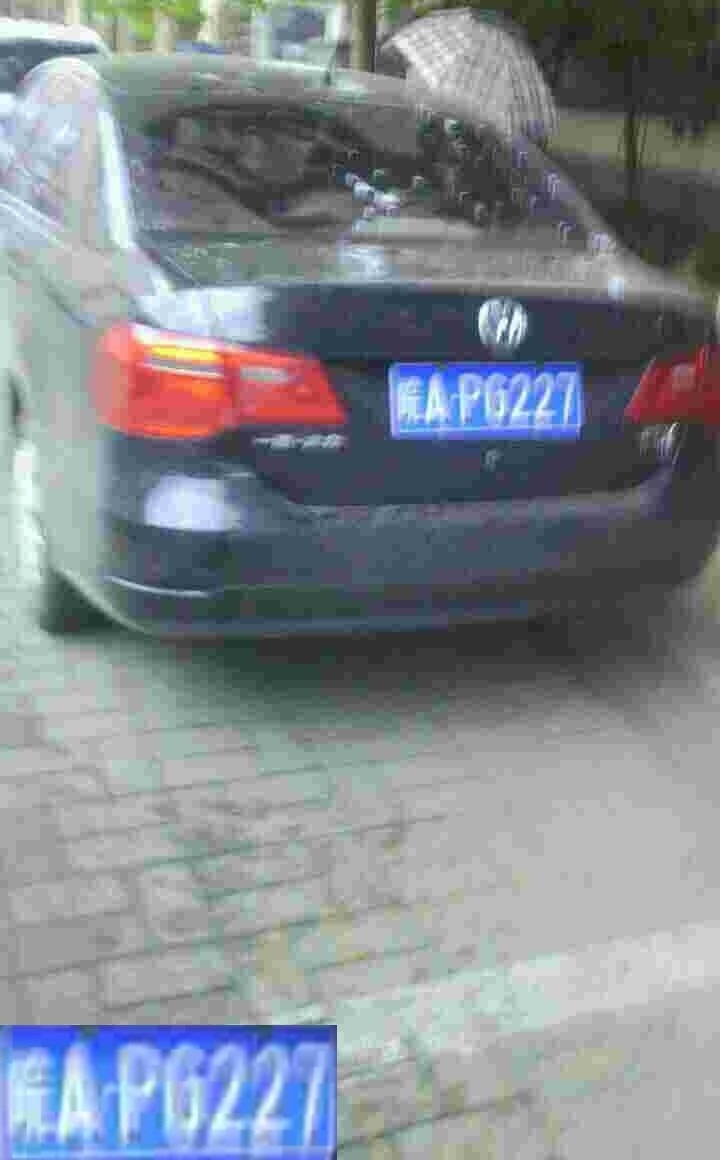}
    }
    
    \subfloat[][Subset Challenge]{
    \includegraphics[width=0.24\linewidth]{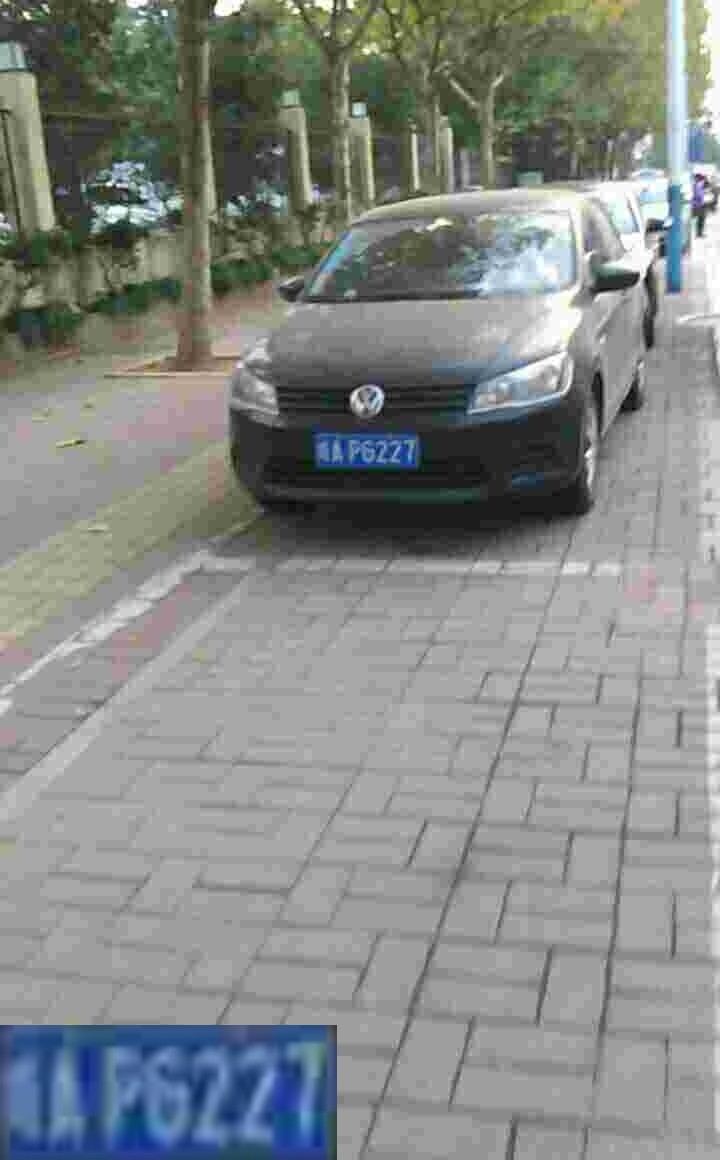}
    }
    
    \subfloat[][Subset Weather]{
    \includegraphics[width=0.24\linewidth]{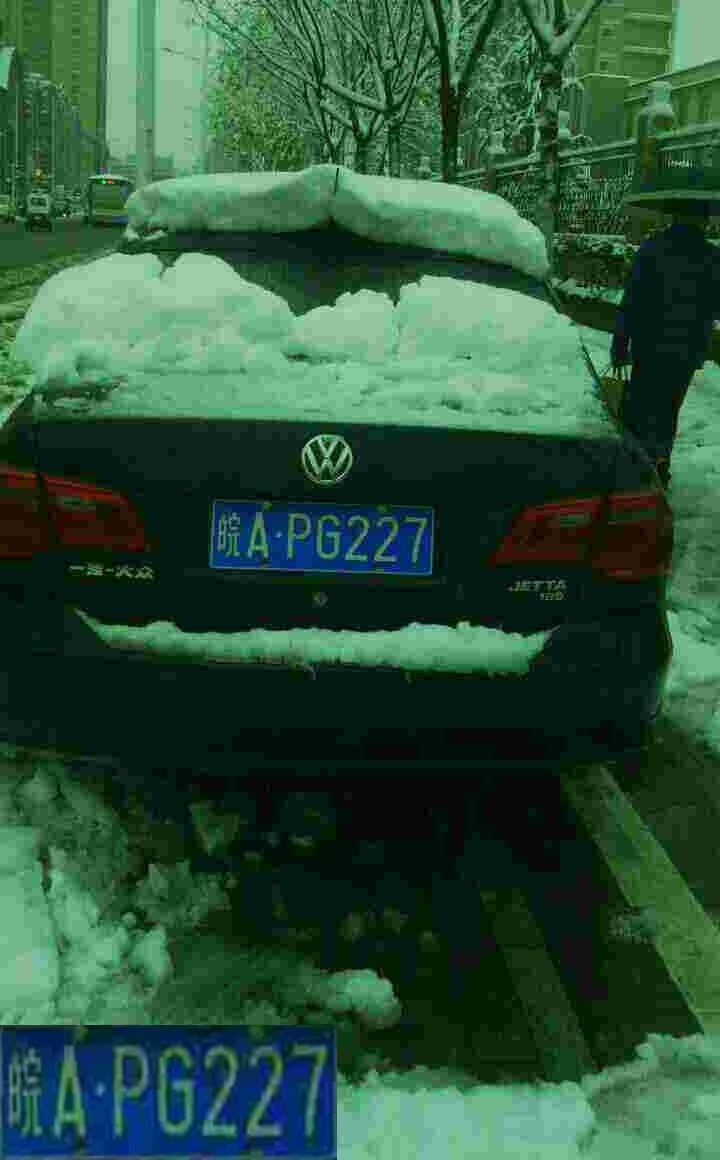}
    }
    
    \subfloat[][Subset Weather]{
    \includegraphics[width=0.24\linewidth]{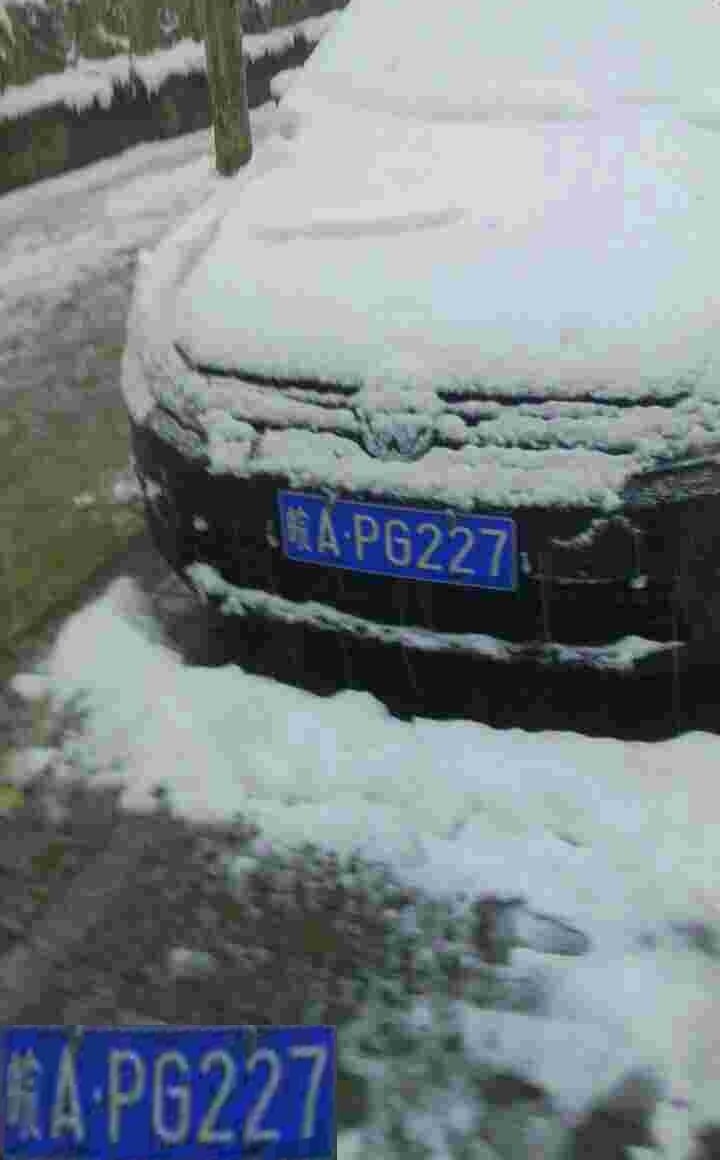}
    }
    }
    }

    \vspace{-0.8mm}
    
    \caption{
    The same vehicle/\gls*{lp} may appear in both training and test images in the \ccpd dataset~\cite{xu2018towards}.
    We show a zoomed-in version of the rectified \gls*{lp} in the lower left region of each image for better viewing.}
    \label{fig:same-vehicle-examples-ccpd}
\end{figure}

The problem with these split protocols is that they do not account for the same vehicle/\gls*{lp} appearing in multiple images, including images from different subsets, as shown in \cref{fig:same-vehicle-examples-aolp} and \cref{fig:same-vehicle-examples-ccpd}.
While one may claim that such images have enough variety to be used both for training and testing \gls*{lp} detectors, as they are fed the entire images, not just the \gls*{lp} region, it seems reasonable to consider that such images should not be employed in the same way (i.e., for both training and testing) in the recognition stage, as the \glspl*{lp} look very similar after being cropped and~rectified.
In fact, they can look very similar even without rectification (e.g., see (d) and (e) in \cref{fig:same-vehicle-examples-aolp}).

\figTeaser shows some examples of near-duplicates from the \aolp and \ccpd datasets, where we picked the $10$th, $50$th, and $90$th percentile image pair for each dataset, according to their Euclidean distance in pixel space.
We found that this metric works reasonably well for this purpose, as images containing the same vehicle are already filtered by the \gls*{lp}~text.

In the \aolp dataset, considering the \emph{AOLP-A} split protocol\customfootnote{We replicated the division made in~\cite{laroca2021efficient,laroca2022cross} of the dataset's images into training, validation and test sets using the split files provided by the authors.}, there are $320$ duplicates from the test set in the training one.
As there are $683$ test images in this protocol, $\textbf{46.9}$\textbf{\%} of them have duplicates.
Startlingly, the number of duplicates is even higher in the \emph{AOLP-B} split protocol, where $413$ of the $611$ test images~($\textbf{67.6}$\textbf{\%}\hspace{-0.4mm}) have duplicates in the training set.

The situation is less severe --~albeit still concerning~-- for the \ccpd dataset, where we found $29{,}943$ duplicates from the test set in the training set.
Despite the much higher number of duplicates in absolute terms, \ccpd's current version has ${\approx}157$K images with labeled \glspl*{lp} in the test set; that is, the duplicates amount to $\textbf{19.1}$\textbf{\%} of the test~images.

\section{Experiments}
\label{sec:experiments}

This section presents the experiments conducted for this work.
We first describe the duplicate-free splits we propose for the \aolp and \ccpd datasets.
Then, we provide a list of the \gls*{ocr} models explored in our assessments, detailing the framework used to implement them as well as relevant hyperparameters.
Afterward, we briefly describe the process of creating synthetic images to avoid overfitting in training the models.
Finally, we report and discuss the results~obtained.

\subsection{Duplicate-Free Splits for the AOLP and CCPD Datasets}
\label{sec:experiments-fair-datasets}

As the \aolp and \ccpd datasets do not have data scraped from the internet (as CIFAR-10 and CIFAR-100 do, for example), we cannot replace the duplicates with new images due to the risk of selection bias or domain shift~\cite{torralba2011unbiased,tommasi2017deeper,barz2020do,laroca2022first}.
Therefore, we present \emph{\textbf{fair splits}} for each dataset where there are no duplicates of the test images in the training set.
As detailed next, we attempted to preserve the key characteristics of the original splits in the new ones as much as~possible.

The \emph{AOLP-Fair-A} split was created as follows.
Following previous works~\cite{xie2018new,laroca2021efficient,liang2022egsanet}, we randomly divided the images of the \aolp dataset into training and test sets with a $2$:$1$ ratio.
Nevertheless, we ensured that distinct images showing the same vehicle/\gls*{lp} (as those shown in \cref{fig:same-vehicle-examples-aolp}) were all in the same set.
Afterward, we allocated $20$\% of the training images for validation.
In this way, the \mbox{AOLP-A} (adopted in previous works) and AOLP-Fair-A protocols have the same number of images for training, testing and~validation.

The core idea of the AOLP-B protocol is to train the approaches on the \gls*{ac} and \gls*{le} subsets and test them on the \gls*{rp} subset~\cite{fan2022improving,qin2022towards,nguyen2022efficient}.
Thus, we created the \emph{AOLP-Fair-B} protocol in the following way.
We kept the original training and validation sets and removed the duplicates from the test set; otherwise, one could ask whether a potential drop in recognition rate is solely due to the reduction in the number of training examples available~\cite{barz2020do}.
In other words, the test sets for the AOLP-B and AOLP-B-Fair splits are different, with the AOLP-B-Fair's test set being a duplicate-free subset of the AOLP-B's test set.
However, the training and validation sets are exactly the same in both~splits.

As mentioned in \cref{sec:datasets-duplicates}, \ccpd's standard split randomly divides the $200$K images of the Base subset into training ($100$K) and validation ($100$K) sets.
All images from the other subsets are used for testing (except Green, which was introduced later and has its own split).
In order to maintain such distribution, we created the \emph{CCPD-Fair} split as follows.
The Base subset was divided into training and validation sets with $100$K images each, as in the original split.
Nevertheless, instead of making this division completely random, we made the training set free of duplicates by allocating all duplicates to the validation set\customfootnote{We trained the \gls*{ocr} models with and without duplicates in CCPD-Fair's validation set, which is used for early stopping and choosing the best weights.
As the results achieved in the test set were essentially the same, we kept the same number of validation images ($100$K-$103$K) as in the original division.
}.
Similarly, we followed the original split for the Green subset as closely as possible, just reallocating the duplicates from the training set to the validation set.
The test set has not changed.
In essence, the original and CCPD-Fair splits use the same $\approx157$K images for testing but have different images in the training and validation sets (each with $\approx103$K images -- about $100$K from Base and $3$K from~Green).

\subsection{OCR models}
\label{sec:experiments-setup}

In this work, we explore six deep learning-based \gls*{ocr} models widely adopted in the literature~\cite{xu2020what,atienza2021data,laroca2022cross}.
Three of them are multi-task networks proposed specifically for \gls*{lpr}: \cnng~\cite{fan2022improving}, \holistic~\cite{spanhel2017holistic} and \multitask\cite{goncalves2018realtime}, while the other three are \gls*{ctc}-, attention- and Transformer-based networks originally proposed for scene text recognition: \starnet~\cite{liu2016starnet}, \trba~\cite{baek2019what}  and \vitstrbase~\cite{atienza2021vitstr},~respectively.

Following~\cite{goncalves2018realtime,goncalves2019multitask,laroca2022cross}, we implemented the multi-task models using Keras (TensorFlow backend).
\ifieee
For the models originally proposed for scene text recognition, following~\cite{atienza2021data,atienza2021vitstr,laroca2022cross}, we used a fork\customfootnote{\hspace{0.2mm}https://github.com/roatienza/deep-text-recognition-benchmark/} (PyTorch) of the repository used to record the first place of ICDAR2013 focused scene text and ICDAR2019~ArT and third place of ICDAR2017 COCO-Text and ICDAR2019 ReCTS~(task1)~\cite{baek2019what}.
\else
For the models originally proposed for scene text recognition, following~\cite{atienza2021data,atienza2021vitstr,laroca2022cross}, we used a fork\customfootnote{\hspace{0.2mm}\href{https://github.com/roatienza/deep-text-recognition-benchmark/}{https://github.com/roatienza/deep-text-recognition-benchmark/}} (PyTorch) of the repository used to record the first place of ICDAR2013 focused scene text and ICDAR2019~ArT and third place of ICDAR2017 COCO-Text and ICDAR2019 ReCTS~(task1)~\cite{baek2019what}.
\fi

Here we list the hyperparameters used in each framework for training the \gls*{ocr} models.
These hyperparameters were determined through experiments on the validation set.
In Keras, we used the Adam optimizer, initial learning rate~=~$10$\textsuperscript{-$3$} (with \textit{ReduceLROnPlateau}'s patience = $3$ and factor~=~$10$\textsuperscript{-$1$}), batch size~=~$64$, max epochs~=~$100$, and patience~=~$7$.
Patience refers to the number of epochs with no improvement, after which decay is applied or training is stopped.
In PyTorch, we adopted the following parameters: Adadelta optimizer, whose decay rate is set to $\rho= 0.99$, $300$K iterations, and batch size~=~$128$.
All experiments were performed on a computer with an AMD Ryzen Threadripper $1920$X $3.5$GHz CPU, $96$~GB of RAM, and an NVIDIA Quadro RTX~$8000$ GPU~($48$~GB).

\subsection{Data Augmentation}
\label{sec:experiments-data-aug}

It is well-known that (i)~\gls*{lpr} datasets usually have a significant imbalance in terms of character classes as a result of \gls*{lp} assignment policies~\cite{goncalves2018realtime,laroca2021efficient,fan2022improving} and (ii)~\gls*{ocr} models are prone to memorize patterns seen in the training stage~\cite{zeni2020weakly,garcia2022out,laroca2022cross};
this phenomenon was termed \emph{vocabulary reliance} in~\cite{wan2020vocabulary}.
To prevent overfitting, we generated many synthetic \gls*{lp} images to improve the training of the recognition~models.

\begin{figure}[!htb]
    \centering

    \resizebox{0.9\linewidth}{!}{
    \includegraphics[width=0.3\linewidth]{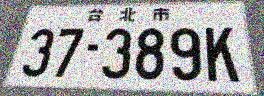}
    \includegraphics[width=0.3\linewidth]{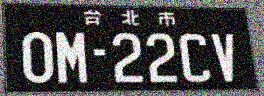}
    \includegraphics[width=0.3\linewidth]{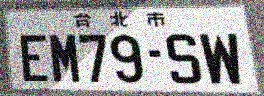}
    }

    \vspace{1mm}

    \resizebox{0.9\linewidth}{!}{
    \includegraphics[width=0.3\linewidth]{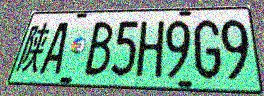}
    \includegraphics[width=0.3\linewidth]{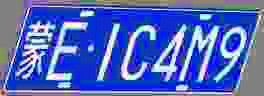}
    \includegraphics[width=0.3\linewidth]{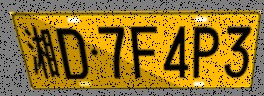}
    }

    \vspace{-2mm}
    
    \caption{Some of the many \gls*{lp} images we created to avoid overfitting.
    The images in the top row simulate \glspl*{lp} from vehicles registered in the Taiwan region (as in \aolp), while those in the bottom row simulate \glspl*{lp} from vehicles registered in mainland~China (as in~\ccpd).
    }
    \label{fig:data-augmentation-samples}
\end{figure}

We created the synthetic \gls*{lp} images as follows.
First, we obtained blank templates that matched the aspect ratio and color scheme of real \glspl*{lp} from the internet.
Then, we superimposed a sequence of characters --~that, although random, mimics the patterns seen on real \glspl*{lp}~-- on each template.
Lastly, we applied various transformations to the \gls*{lp} images to increase variability.
Transformations applied include, but are not limited to, random perspective transform, random noise, random shadows, and random perturbations of hue, saturation and brightness (note that these same transformations were also applied to real training images as a form of data augmentation).
Examples of synthetic \gls*{lp} images generated in this way can be seen in \cref{fig:data-augmentation-samples}.

\subsection{Results}
\label{sec:experiments-results}

Here, we report the recognition rates reached by the \gls*{ocr} models in each dataset under the original and fair splits\customfootnote{We reinforce that all results reported in this work (Table \ref{tab:results-aolp-a} to \ref{tab:results-ccpd-detailed}) are from our experiments (i.e., we trained all \gls*{ocr} models following precisely the same protocol in each set of experiments) and not replicated from the cited~papers.}.
Recognition rate refers to the number of correctly recognized \glspl*{lp} divided by the number of \glspl*{lp} in the test set~\cite{laroca2021efficient,silva2022flexible,wang2022rethinking}.
Following~\cite{barz2020do}, in addition to the recognition rates obtained in the original and fair protocols, we report their differences in terms of absolute percentage points (``Gap'') and in relation to the original error (``Rel.~Gap'') $(gap/(100\%-acc))$.

\RLI{ver sobre adicionar sistemas comerciais}

The results of all \gls*{ocr} models on the \aolp dataset are shown in \cref{tab:results-aolp-a} and \cref{tab:results-aolp-b}.
In both protocols (AOLP-A and AOLP-B), the recognition rates obtained in the fair split were considerably lower than those achieved in the original one.
Specifically, \emph{the error rates were more than twice as high in the experiments conducted under the fair~protocols}.

\begin{table}[!htb]
\centering
\caption{Recognition rates achieved by six \gls*{ocr} models under the AOLP-A (adopted in previous works) and AOLP-Fair-A (ours) protocols. The best value in each column is shown in bold.}
\label{tab:results-aolp-a}

\vspace{-2mm}

\resizebox{0.975\linewidth}{!}{
\begin{tabular}{lcccc}
\toprule
Model        & AOLP-A $\uparrow$    & AOLP-A-Fair $\uparrow$ & Gap $\downarrow$ & Rel. Gap $\downarrow$ \\
\midrule
\cnng~\cite{fan2022improving}         & $98.88$\% & $95.63$\% & $3.25$\% & $290.2$\%   \\
\holistic~\cite{spanhel2017holistic} & $96.75$\% & $93.11$\% & $3.64$\% & $\textbf{112.0}$\textbf{\%}  \\
\multitask~\cite{goncalves2018realtime}   & $97.33$\% & $93.79$\% & $3.54$\% & $132.6$\%  \\
\starnet~\cite{liu2016starnet}     & $98.69$\% & $95.83$\% & $2.86$\% & $218.3$\%   \\
\trba~\cite{baek2019what}         & $\textbf{99.18}$\textbf{\%} & $\textbf{96.94}$\textbf{\%} & $2.24$\% & $273.2$\%   \\
\vitstrbase~\cite{atienza2021vitstr}  & $98.74$\% & $\textbf{96.94}$\textbf{\%} & $\textbf{1.80}$\textbf{\%} & $142.9$\%   \\ \bottomrule
\end{tabular}
}
\end{table}

\begin{table}[!htb]
\centering
\caption{Recognition rates achieved by six \gls*{ocr} models under the AOLP-B (adopted in previous works) and AOLP-Fair-B (ours) protocols. The best value in each column is shown in bold.}
\label{tab:results-aolp-b}

\vspace{-2mm}

\resizebox{0.975\linewidth}{!}{
\begin{tabular}{lcccc}
\toprule
Model        & AOLP-B $\uparrow$    & AOLP-B-Fair $\uparrow$ & Gap $\downarrow$ & Rel. Gap $\downarrow$ \\
\midrule
\cnng~\cite{fan2022improving}         & $\textbf{98.91}$\textbf{\%} & $96.80$\% & $2.11$\% & $193.6$\%   \\
\holistic~\cite{spanhel2017holistic} & $98.42$\% & $96.30$\% & $2.12$\% & $134.2$\%  \\
\multitask~\cite{goncalves2018realtime}   & $98.42$\% & $95.29$\% & $3.13$\% & $198.1$\%  \\
\starnet~\cite{liu2016starnet}     & $98.47$\% & $96.46$\% & $2.01$\% & $131.4$\%   \\
\trba~\cite{baek2019what}         & $98.75$\% & $\textbf{97.47}$\textbf{\%} & $\textbf{1.28}$\textbf{\%} & $\textbf{102.4}$\textbf{\%}   \\
\vitstrbase~\cite{atienza2021vitstr}  & $98.75$\% & $97.31$\% & $1.44$\% & $115.2$\%   \\ \bottomrule
\end{tabular}
}
\end{table}

\RLI{procurar imagens que todos os modelos acertaram no trad, mas que todos erraram no fair (provavelmente ache algo interessante)}

It is crucial to note that the ranking of \gls*{ocr} models \emph{changed} when they were trained and tested under fair splits.
For example, the \cnng model achieved the best result under the AOLP-B protocol (as in~\cite{fan2022improving}, where it was proposed) but only reached the third-best result under AOLP-Fair-B.
Similarly, the ViTSTR-Base model ranked third under the AOLP-A protocol but tied for first place with \trba under~AOLP\nobreakdash-Fair\nobreakdash-A.

These results strongly suggest that, in the past, the high fraction of near-duplicates in the splits traditionally adopted in the literature for the \aolp dataset may have prevented the publication and adoption of more efficient \gls*{lpr} models that can generalize as well as other models but fail to memorize duplicates.
A similar concern was raised by Barz et al.~\cite{barz2020do} with respect to the CIFAR-10 and CIFAR-100 datasets.

\cref{tab:results-ccpd} shows the results for the \ccpd dataset.
\cref{tab:results-ccpd-detailed} breaks down the results for each of the \ccpd's subsets, as is commonly done in the literature~\cite{xu2018towards,zhang2021robust_attentional,wang2022rethinking}.
While the largest drop in recognition rate was $3.64$\% in the \aolp dataset, the \starnet and \trba models had drops of $5.20$\% and $4.35$\% in recognition rate under the CCPD-Fair protocol, respectively.
The average recognition rate decreased from $80.3$\% to $77.6$\%, with the relative gaps being much smaller than those observed in the \aolp dataset because the recognition rates reached in \ccpd were not as high (this was expected, as the authors of the \ccpd dataset modified it twice with the specific purpose of making it much more challenging than it was~initially).

\begin{table}[!htb]
\centering
\caption{Recognition rates achieved by six well-known \gls*{ocr} models on the \ccpd dataset under the standard and CCPD-Fair protocols. The best value in each column is shown in bold.}
\label{tab:results-ccpd}

\vspace{-2mm}

\resizebox{0.975\linewidth}{!}{
\begin{tabular}{lcccc}
\toprule
Model        & \ccpd\ $\uparrow$    & \ccpdfair $\uparrow$ & Gap $\downarrow$ & Rel. Gap $\downarrow$ \\
\midrule
\cnng~\cite{fan2022improving}         & $\textbf{88.24}$\textbf{\%} & $\textbf{86.93}$\textbf{\%} & $1.31$\% & $11.1$\%   \\
\holistic~\cite{spanhel2017holistic} & $77.01$\% & $75.41$\% & $1.60$\% & $\phantom{0}7.0$\%  \\
\multitask~\cite{goncalves2018realtime}   & $83.01$\% & $81.84$\% & $\textbf{1.17}$\textbf{\%} & $\phantom{0}\textbf{6.9}$\textbf{\%}  \\
\starnet~\cite{liu2016starnet}     & $78.53$\% & $73.33$\% & $5.20$\% & $24.2$\%   \\
\trba~\cite{baek2019what}         & $75.83$\% & $71.48$\% & $4.35$\% & $18.0$\%   \\
\vitstrbase~\cite{atienza2021vitstr}  & $79.06$\% & $76.37$\% & $2.69$\% & $12.9$\%   \\ \bottomrule
\end{tabular}
}
\end{table}

\begin{table}[!htb]
\centering
\setlength{\tabcolsep}{3pt}
\caption{
Recognition rates (\%) achieved on each subset of the \ccpd dataset under the standard and CCPD-Fair protocols.
}
\label{tab:results-ccpd-detailed}

\vspace{-2mm}

\resizebox{0.99\linewidth}{!}{
\begin{tabular}{@{}lccccccccc@{}}
\toprule
\multirow{2}{*}{\diagbox[trim=l,innerrightsep=7.75pt]{Model}{Subset}}        & \multirow{2}{*}{\begin{tabular}[c]{@{}c@{}}Blur\\21K\end{tabular}}   & \multirow{2}{*}{\begin{tabular}[c]{@{}c@{}}Chal.\\$50$K\end{tabular}} & \multirow{2}{*}{\begin{tabular}[c]{@{}c@{}}DB\\$10$K\end{tabular}}     & \multirow{2}{*}{\begin{tabular}[c]{@{}c@{}}FN\\$21$K\end{tabular}}     & \multirow{2}{*}{\begin{tabular}[c]{@{}c@{}}Green\\$5$K\end{tabular}}  & \multirow{2}{*}{\begin{tabular}[c]{@{}c@{}}Rot.\\$10$K\end{tabular}} & \multirow{2}{*}{\begin{tabular}[c]{@{}c@{}}Tilt\\$30$K\end{tabular}}   & \multirow{2}{*}{\begin{tabular}[c]{@{}c@{}}Weath.\\$10$K\end{tabular}} & \multirow{2}{*}{\begin{tabular}[c]{@{}c@{}}\phantom{i}All\phantom{i}\\$157$K\end{tabular}}  \\
& & & & & & & & & \\ \midrule
\textbf{\textit{\ccpd}} & & & & & & & & & \\
\customIndent~\cnng~\cite{fan2022improving}         & $\phantom{0}77.3\phantom{0}$ & $\phantom{0}84.1\phantom{0}$    & $\phantom{0}80.8\phantom{0}$ & $\phantom{0}91.0\phantom{0}$ & $\phantom{0}94.2\phantom{0}$ & $\phantom{0}97.4\phantom{0}$ & $\phantom{0}95.5\phantom{0}$ & $\phantom{0}99.3\phantom{0}$  & $\textbf{88.2}$ \\
\customIndent~\holistic~\cite{spanhel2017holistic}\phantom{i} & $52.0$ & $68.8$    & $67.8$ & $81.9$ & $93.0$ & $95.2$ & $91.4$ & $99.1$  & $77.0$ \\
\customIndent~\multitask~\cite{goncalves2018realtime}   & $68.4$ & $77.1$    & $73.2$ & $86.1$ & $93.8$ & $96.0$ & $92.6$ & $98.8$  & $83.0$ \\
\customIndent~\starnet~\cite{liu2016starnet}     & $58.7$ & $71.2$    & $64.9$ & $83.3$ & $91.7$ & $94.9$ & $91.2$ & $98.4$  & $78.5$ \\
\customIndent~\trba~\cite{baek2019what}         & $50.2$ & $67.9$    & $59.6$ & $81.9$ & $92.7$ & $94.7$ & $91.1$ & $98.4$  & $75.8$ \\
\customIndent~\vitstrbase~\cite{atienza2021vitstr}  & $56.4$ & $72.0$    & $65.9$ & $84.6$ & $94.0$ & $95.5$ & $92.2$ & $98.8$  & $79.1$ \\ \midrule
\textbf{\textit{\ccpdfair}} & & & & & & & & & \\
\customIndent~\cnng~\cite{fan2022improving}         & $73.4$ & $82.8$    & $78.8$ & $90.2$ & $92.8$ & $97.0$ & $95.1$ & $99.2$  & $\textbf{86.9}$ \\
\customIndent~\holistic~\cite{spanhel2017holistic} & $47.9$ & $66.8$    & $65.6$ & $81.2$ & $91.2$ & $95.1$ & $90.9$ & $98.2$  & $75.4$ \\
\customIndent~\multitask~\cite{goncalves2018realtime}   & $65.7$ & $75.7$    & $71.5$ & $85.3$ & $92.0$ & $95.6$ & $92.2$ & $98.7$  & $81.8$ \\
\customIndent~\starnet~\cite{liu2016starnet}     & $46.4$ & $64.3$    & $57.2$ & $79.7$ & $91.5$ & $93.9$ & $89.6$ & $98.0$  & $73.3$ \\
\customIndent~\trba~\cite{baek2019what}         & $38.7$ & $62.7$    & $52.4$ & $80.0$ & $91.2$ & $93.8$ & $89.3$ & $98.1$  & $71.5$ \\
\customIndent~\vitstrbase~\cite{atienza2021vitstr}  & $50.2$ & $68.4$    & $63.5$ & $82.5$ & $93.5$ & $95.1$ & $91.1$ & $98.7$  & $76.4$ \\
\bottomrule
\end{tabular}
}
\end{table}

\RLI{como conjunto de dados CCPD possui muitas imagens, apesar do GAP ser pequeno (em termos de pontos percentuais), são muitos erros em números absolutos no protocolo Fair. No melhor caso, 1131 erros a mais na multitask e no pior caso 8153 na starnet.}

Examining the absolute number of errors may give a clearer understanding of the impact of duplicates on the evaluation of the recognition models.
The lowest performance gap of $1.17$\% translates to $1{,}800$+ additional \glspl*{lp} being misrecognized under the fair split (vs. the standard one), while the highest performance gap of $5.2$\% represents a staggering number of $8{,}000$+ more \glspl*{lp} being incorrectly recognized under the fair~split.

Differently from the results obtained in the \aolp dataset, the ranking of models remained practically the same in \ccpd; only the fourth and fifth places switched positions.
This is partially due to the significant performance gap between the models and suggests that the community's research efforts have not \emph{yet} overfitted to the presence of duplicates in the standard split of the \ccpd dataset.
However, we fundamentally believe it is only a matter of time before this starts to happen or be noticed (potentially with the use of deeper models, as the ability to memorize training data increases with the model's capacity~\cite{barz2020do,hooker2020characterising}) in case such near-duplicates in the training and test sets are not acknowledged and therefore~avoided.

\RLI{mean confidence trad vs fair; maior confianca permite que regras heuristicas sejam usadas... ou seja, impacto pode ser ainda maior do que o observado nestes experimentos}

\section{What about other datasets?}
\label{sec:other-datasets}

As mentioned earlier, we focused our analysis on the \aolp and \ccpd datasets due to their predominance in the \gls*{alpr} literature~\cite{xie2018new,qin2020efficient,zhang2020robust_license,liang2022egsanet,pham2022effective}.
Nevertheless, as this issue (i.e., \gls*{lpr} models being evaluated in datasets containing near-duplicates in the training and test sets) has not yet received due attention from the community, it has recurred in assessments carried out on several other public~datasets.

\begin{figure}[!htb]
    \captionsetup[subfigure]{labelformat=empty}
    \centering

    \resizebox{0.99\linewidth}{!}{
    \subfloat[]{
    \includegraphics[height=13ex]{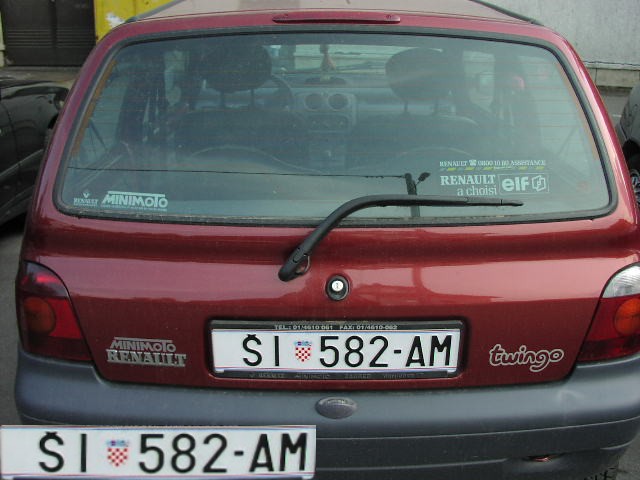}
    } \hspace{-2.25mm}
    \subfloat[]{
    \includegraphics[height=13ex]{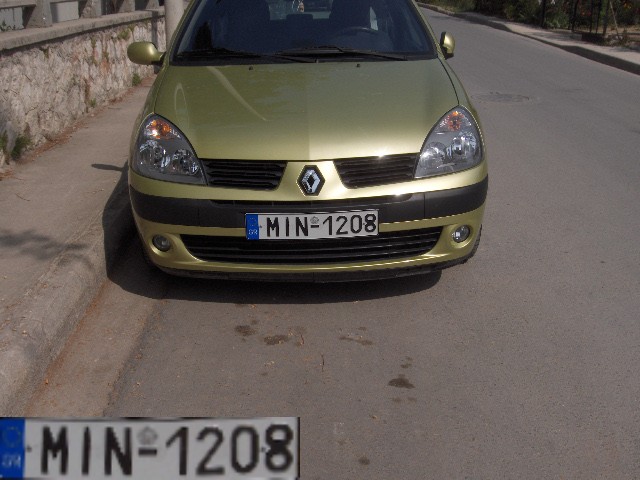}
    } \hspace{-2.25mm}
    \subfloat[]{
    \includegraphics[height=13ex]{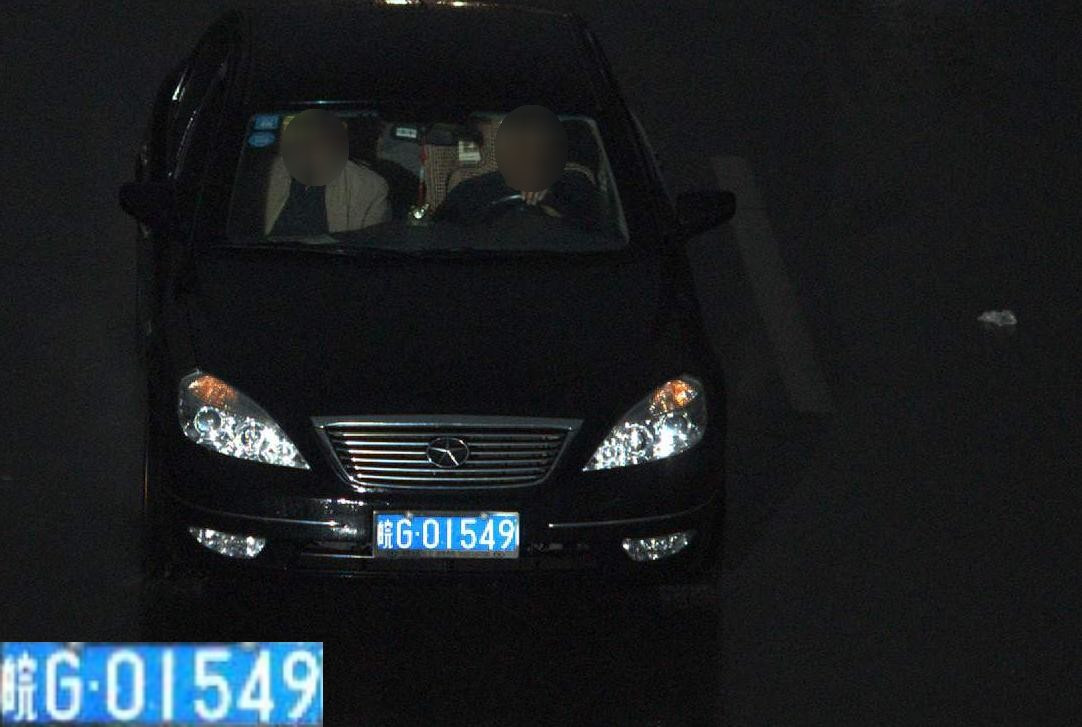}
    } \,
    } 

    \vspace{-3.1mm}

    \resizebox{0.99\linewidth}{!}{
    \subfloat[(a) \englishlp~\cite{englishlp}]{
    \includegraphics[height=13ex]{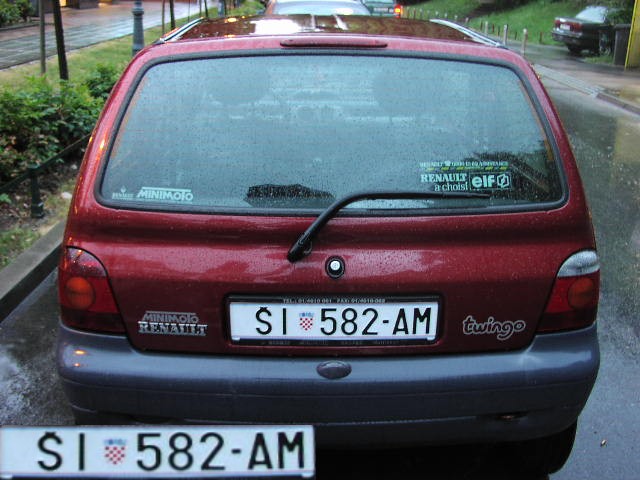}
    } \hspace{-2.25mm}
    \subfloat[(b) \medialab~\cite{anagnostopoulos2008license}]{
    \includegraphics[height=13ex]{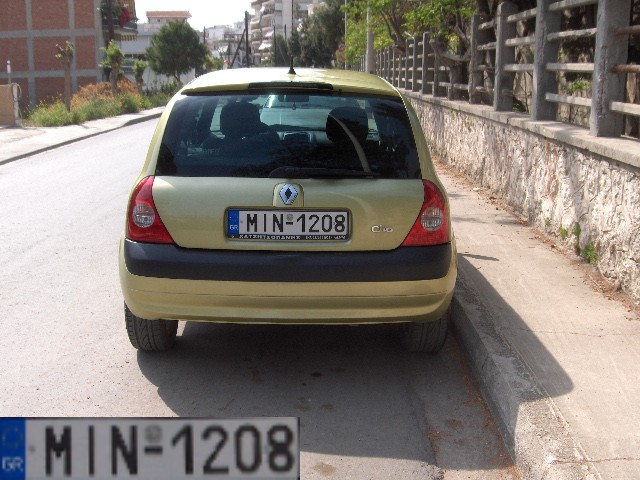}
    } \hspace{-2.25mm}
    \subfloat[(c) \pku~\cite{yuan2017robust}]{
    \includegraphics[height=13ex]{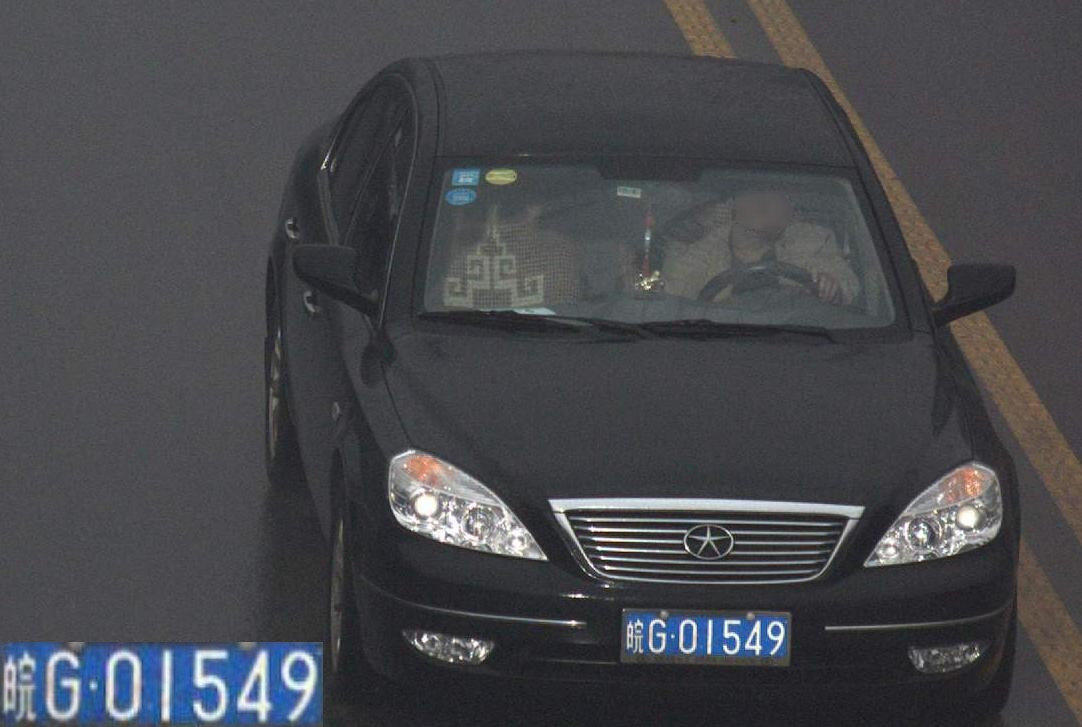}
    } \,
    }
    
    \caption{
    \gls*{alpr} datasets that do not have a well-defined evaluation protocol are customarily divided into training and test sets randomly without the authors noticing that the same vehicle/\gls*{lp} may appear in multiple images.
    Above, we show a pair of near-duplicates from each of the \englishlp, \pku and \medialab datasets.
    Observe that it is common for an \gls*{lp} to look very similar in different images even without rectification.
    We show a zoomed-in version of the rectified \gls*{lp} in the lower left region of each image for better~viewing.}
    \label{fig:duplicates-other-datasets}
\end{figure}

Consider the \englishlp~\cite{englishlp}, \medialab~\cite{anagnostopoulos2008license} and \pku~\cite{yuan2017robust} datasets as examples (they are quite popular, albeit far less than \aolp and \ccpd).
They all have near-duplicates, as shown in \cref{fig:duplicates-other-datasets}.
As these datasets lack an official evaluation protocol, it is common for authors to divide their images into training, validation and test sets randomly~\cite{zhuang2018towards,gao2020edflpr,khan2021performance,laroca2021efficient,zhang2021efficient,qin2022towards}.
As can be inferred, the presence of near-duplicates in these datasets has also been overlooked in such~setups.

\begin{figure}[!htb]
    \centering
    \captionsetup[subfigure]{skip=-1pt}

    \resizebox{0.99\linewidth}{!}{
    \subfloat[Training set]{
    \includegraphics[width=0.31\linewidth]{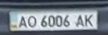}
    \includegraphics[width=0.31\linewidth]{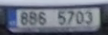}
    \includegraphics[width=0.31\linewidth]{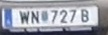}
    } \,
    }

    \vspace{2mm}

    \resizebox{0.99\linewidth}{!}{
    \subfloat[Test set]{
    \includegraphics[width=0.31\linewidth]{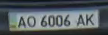}
    \includegraphics[width=0.31\linewidth]{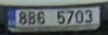}
    \includegraphics[width=0.31\linewidth]{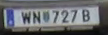} 
    } \,
    }

    \vspace{-0.5mm}
    
    \caption{Examples of near-duplicates in the \reid dataset~\cite{spanhel2017holistic}.
    It is clear that such duplicates may also considerably bias the evaluation of \gls*{alpr} systems that do not perform rectification before the \gls*{lpr} stage (e.g.,~\cite{li2019toward,henry2020multinational,laroca2021efficient}).
    }
    
    \label{fig:duplicates-reid}
\end{figure}

The \reid dataset~\cite{spanhel2017holistic} differs from the datasets mentioned above by having a standard protocol.
It has $182{,}335$ images of cropped low-resolution \glspl*{lp}, of which $105{,}923$ are in the training set and $76{,}412$ are in the test set.
We found that $52{,}394$~($\textbf{68.6}$\textbf{\%}\hspace{-0.4mm}) of the test images have near-duplicates in the training set (see some examples in \cref{fig:duplicates-reid}).
Although alarming, the high fraction of duplicates has gone unacknowledged in works using the \reid dataset for experimentation~\cite{spanhel2018geometric,wu2019pixtextgan,moussa2022forensic}.

\begin{figure}[!htb]
    \centering

    \resizebox{0.99\linewidth}{!}{
    \subfloat[][Images from \chineselp~\cite{zhou2012principal}]{
        \includegraphics[height=13.8ex]{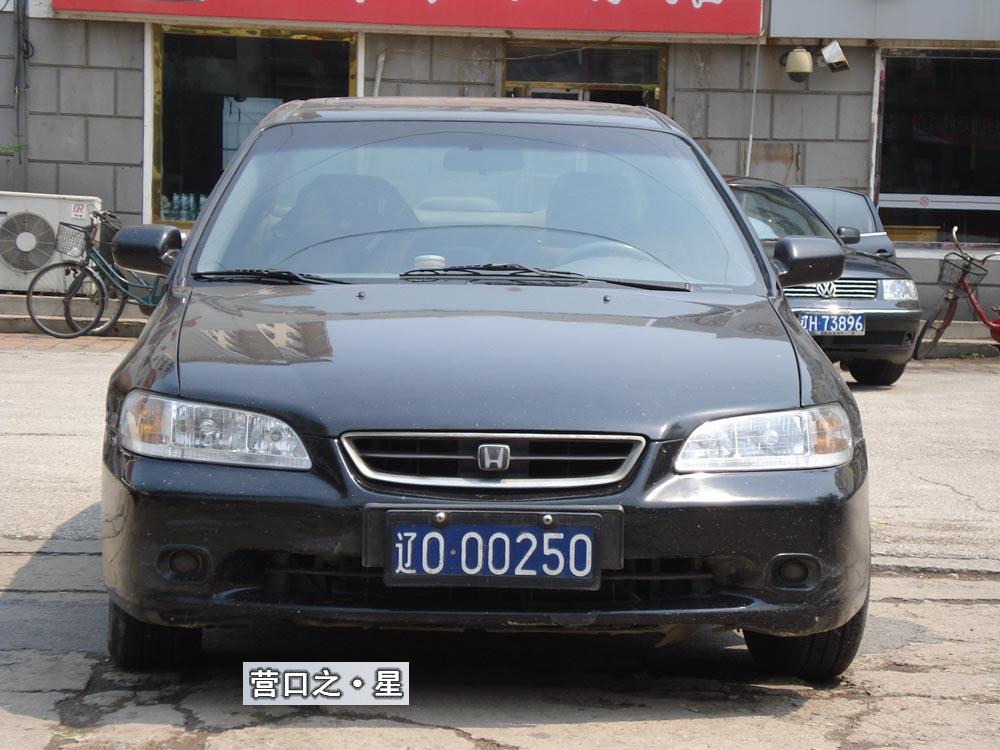}
        \includegraphics[height=13.8ex]{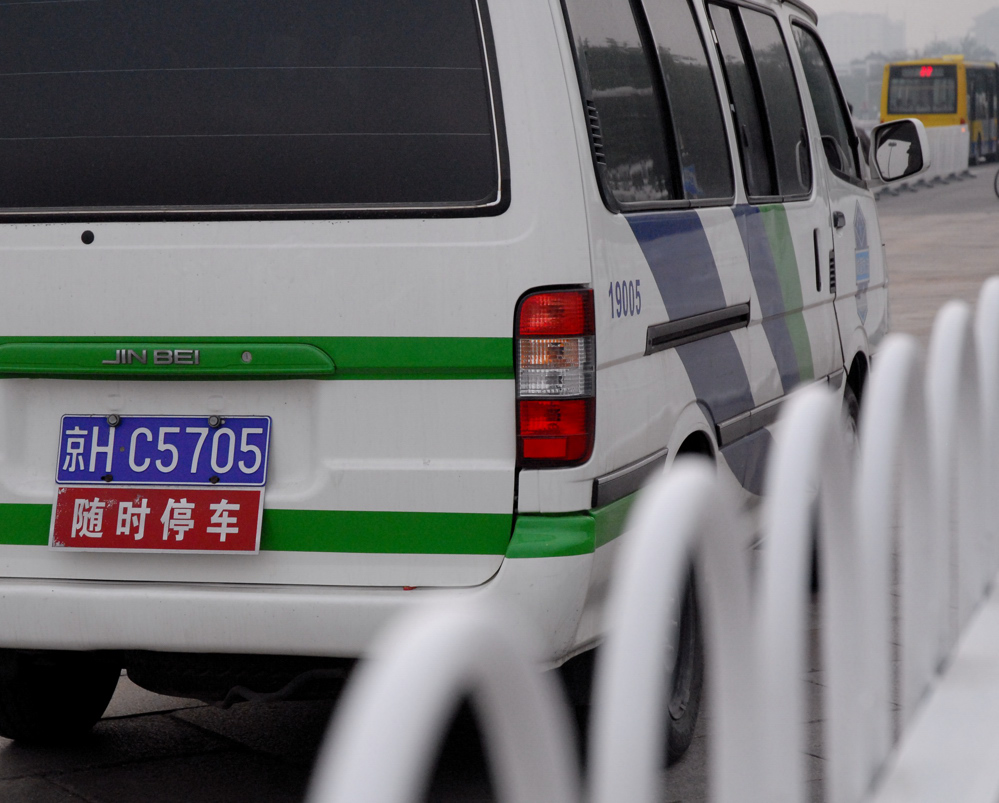}
        \includegraphics[height=13.8ex]{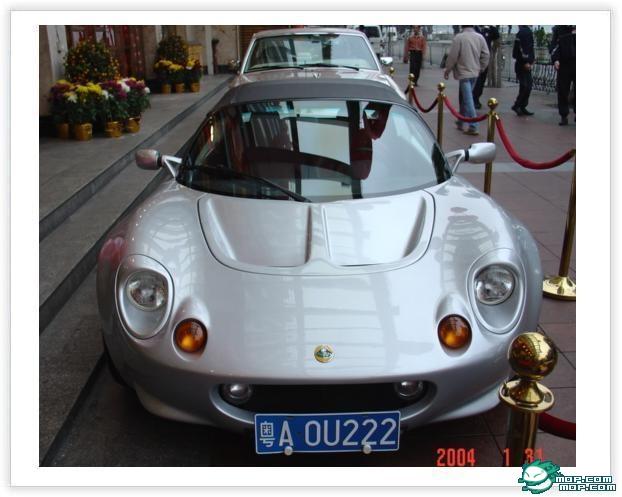} 
    } \,
    }

    \vspace{3mm}
    
    \resizebox{0.99\linewidth}{!}{
    \subfloat[][Images from \clpd~\cite{zhang2021robust_attentional}]{
        \includegraphics[height=15ex]{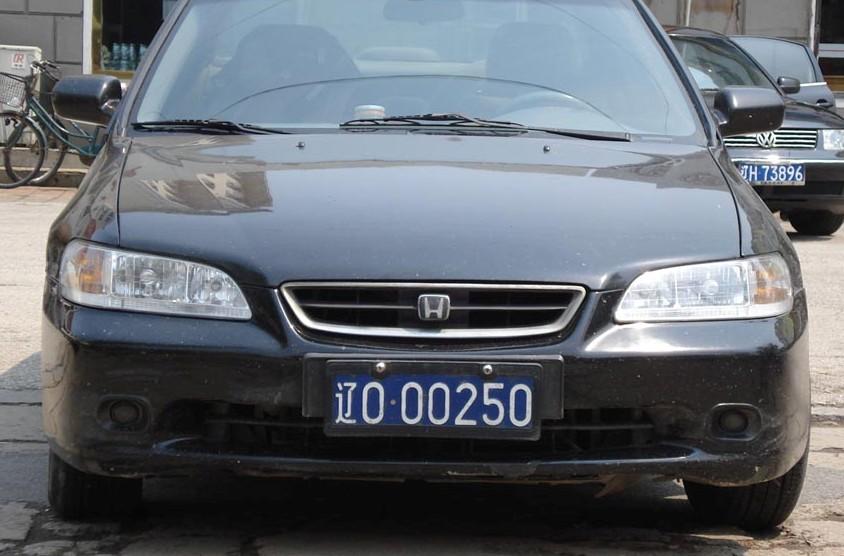}      \includegraphics[height=15ex]{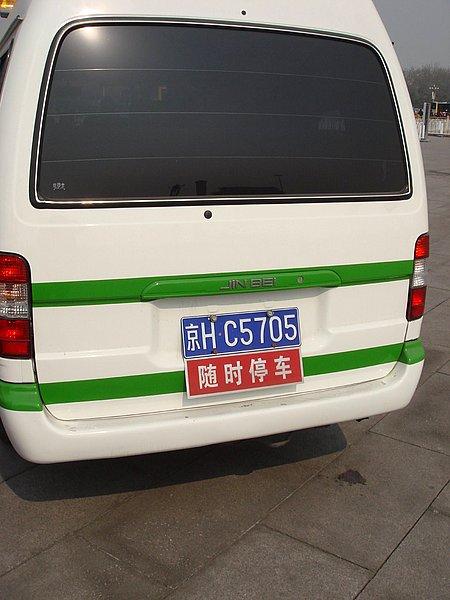}
        \includegraphics[height=15ex]{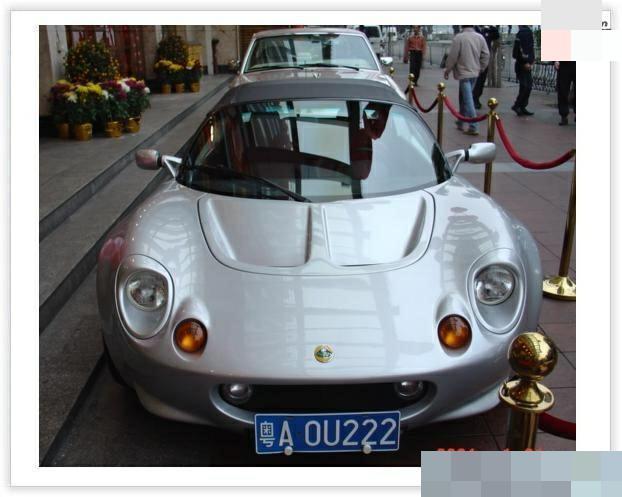}
    } \,
    }

    \caption{There are duplicates even across different datasets.
    The above images were taken from the \chineselp and \clpd datasets, both of which contain images scraped from the internet. 
    The presence of near-duplicates across datasets can significantly bias the results of cross-dataset~experiments.
    }
    \label{fig:same-vehicle-examples-chineselp-and-clpd}
\end{figure}

We also want to draw attention to the fact that there are duplicates even across different datasets.
Recently, Zhang et al.~\cite{zhang2021robust_attentional} released the \clpd dataset, which comprises $1{,}200$ images gathered from multiple sources such as the internet, mobile phones, and car driving recorders.
The authors employed all images for testing to verify the practicality of their \gls*{lpd} and \gls*{lpr} models, trained on other datasets.
Subsequent studies have followed this protocol~\cite{zou2020robust,liu2021fast,zhang2021efficient,fan2022improving,wang2022rethinking}.
The problem is that several vehicles/\glspl*{lp} shown in \clpd are also shown in the \chineselp dataset~\cite{zhou2012principal} (see \cref{fig:same-vehicle-examples-chineselp-and-clpd}).
That is, if not yet, images from the \chineselp dataset will eventually be used to train \gls*{alpr} systems that will then be tested on the \clpd dataset.
These experiments will likely be regarded as ``cross-dataset,'' although perhaps they should~not.

One last example that reinforces how this issue has gone unnoticed in the literature is~\cite{gong2022unified}, where the authors presented a detailed comparison between multiple datasets gathered in mainland China --~including \chineselp and \clpd~-- without noticing the existence of duplicates across~them.

It is essential to acknowledge that there are datasets, albeit very few, where the authors deliberately defined a standard split with no duplicates within the training and test sets.
We verified that RodoSol-ALPR~\cite{laroca2022cross} is one such~dataset.
\section{Conclusions}
\label{sec:conclusions}

We drew attention to the large fraction of near-duplicates in the training and test sets of datasets widely adopted in \gls*{alpr} research.
Both the existence of such duplicates and their influence on the performance evaluation of \gls*{lpr} models have largely gone unnoticed in the~literature.

Our experiments on the \aolp and \ccpd datasets, the most commonly used in the field, showed that the presence of  near-duplicates significantly impacts the performance evaluation of \gls*{ocr} models applied to \gls*{lpr}.
In the \aolp dataset, the error rates reported by the models were more than twice as high in the experiments conducted under the fair splits.
The ranking of models also changed when they were trained and tested under duplicate-free splits.
In the more challenging \ccpd dataset, the models showed recognition rate drops of up to~$5.2$\%.
Specifically, the average recognition rate decreased from $80.3$\% to $77.6$\% when the experiments were conducted under the fair split compared to the standard one.
These results indicate that duplicates have biased the evaluation and development of deep learning-based models for~\gls*{lpr}.

We created the \emph{fair splits} for the abovementioned datasets by dividing their images into new training, validation and test sets while ensuring that no duplicates from the test set are present in the training set and preserving the original splits' key characteristics as much as possible.
These new splits and the list of duplicates found are publicly~available.

We hope this work will encourage \gls*{lpr} researchers to train and evaluate their models using the fair splits we created for the \aolp and \ccpd datasets and to beware of duplicates when performing experiments on other datasets.
This work also provides researchers with a clearer understanding of the true capabilities of LPR models that have only been evaluated on test sets that include duplicates from the training~set.

\ifieee
Further examination of the occurrences of near-duplicates in other \gls*{alpr} datasets, including those mentioned in \cref{sec:other-datasets}, will be conducted in future research.
We also want to perform extra statistical analysis to confirm that the data distributions in the traditionally adopted and fair splits are~compatible.
\else
Further examination of the occurrences of near-duplicates in other \gls*{alpr} datasets, including those mentioned in \cref{sec:other-datasets}, will be conducted in future~research.
\fi

\RLI{avaliação do impacto de tais duplicadsa no estágio de detecção fica para trabalhos futuros.}

\RLI{ou aqui ou na introdução, esse trabalho serve para guiar novos datasets... criadores de novos datasets anteciparem este problema das duplicadas...}

\RLI{por questões de espaço, este trabalho focamos nestes dois conjuntos de dados. mas ja encontramos problemas em outros conjuntos de dados. por exemplos imagens repetidas no conjunto de dados ChineseLP e CLPD}
\section*{\uppercase{Acknowledgments}}

\iffinal
    This work was partly supported by the Coordination for the Improvement of Higher Education Personnel~(CAPES) (\textit{Programa de Coopera\c{c}\~{a}o Acad\^{e}mica em Seguran\c{c}a P\'{u}blica e Ci\^{e}ncias Forenses \#~88881.516265/2020\nobreakdash-01}), and partly by the National Council for Scientific and Technological Development~(CNPq) (\#~308879/2020\nobreakdash-1 and \#~306878/2022\nobreakdash-4).
    We thank the support of NVIDIA Corporation with the donation of the Quadro RTX $8000$ GPU used for this~research.
\else
    The acknowledgments are hidden for review.
\fi

\balance
\bibliographystyle{IEEEtran}
\ifieee
\bibliography{bibtex-short}
\else
\bibliography{bibtex-short}
\fi
\end{document}